\documentclass[11pt]{article}

\usepackage{amsmath,amsfonts,bm}

\def\eqref#1{equation~\ref{#1}}

\def\1{\bm{1}}

\DeclareMathAlphabet{\mathsfit}{\encodingdefault}{\sfdefault}{m}{sl}
\SetMathAlphabet{\mathsfit}{bold}{\encodingdefault}{\sfdefault}{bx}{n}

\usepackage{hunyuan}

\usepackage{xcolor}
\usepackage{booktabs}
\usepackage{makecell}
\usepackage{fancyhdr}
\usepackage{tcolorbox}

\definecolor{hunyuanblue}{HTML}{1E4A8F}
\definecolor{abstractbg}{HTML}{F0F7FC}
\definecolor{boxfill}{HTML}{EEF3FB}
\definecolor{boxframe}{HTML}{1E4A8F}
\definecolor{boxtitle}{HTML}{12305C}
\definecolor{boxtitlebg}{HTML}{B8CEE8}

\newtcolorbox{highlightbox}[1]{
  colback=boxfill, colframe=boxframe, boxrule=0.9pt, arc=3pt,
  left=10pt, right=10pt, top=8pt, bottom=8pt,
  colbacktitle=boxtitlebg, coltitle=boxtitle,
  fonttitle=\bfseries,
  title={#1},
}
\newcommand{\dashrule}{\leaders\hbox to 6pt{\hss\rule[0.4ex]{4pt}{0.8pt}\hss}\hfill\kern0pt}

\hypersetup{
  colorlinks=true,
  linkcolor=hunyuanblue,
  citecolor=hunyuanblue,
  urlcolor=hunyuanblue,
}

\makeatletter
\long\def\@makecaption#1#2{%
  \vskip 10pt
  \setbox\@tempboxa\hbox{#1: #2}%
  \ifdim \wd\@tempboxa >\hsize
    \noindent #1: #2\par
  \else
    \hbox to\hsize{\hfil\box\@tempboxa\hfil}%
  \fi}
\def\section{\@startsiction{section}{1}{\z@}{-0.24in}{0.10in}
             {\large\bf\raggedright\color{hunyuanblue}}}
\def\subsection{\@startsection{subsection}{2}{\z@}{-0.20in}{0.08in}
                {\normalsize\bf\raggedright\color{hunyuanblue}}}
\makeatother

\renewcommand{\headrulewidth}{0pt}
\renewcommand{\footrulewidth}{0pt}

\fancypagestyle{plain}{%
  \fancyhf{}%
  \fancyfoot[C]{\thepage}%
  \renewcommand{\headrulewidth}{0pt}%
  \renewcommand{\footrulewidth}{0pt}%
}

\fancypagestyle{firststyle}{%
  \fancyhf{}%
  \fancyhead[L]{\raisebox{-18.5mm}[0pt][0pt]{\includegraphics[height=12mm]{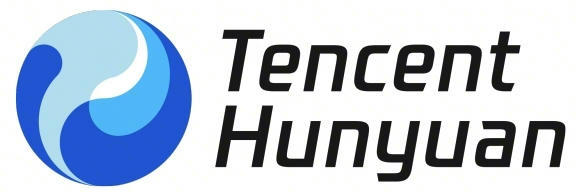}}}%
  \fancyfoot[C]{\thepage}%
  \renewcommand{\headrulewidth}{0pt}%
  \renewcommand{\footrulewidth}{0pt}%
}

\begin{document}

\thispagestyle{firststyle}
\vspace*{0.25cm}
{\color{hunyuanblue}\hrule height 0.6pt}
\vskip 6mm
\begin{center}
{\LARGE\bfseries \raisebox{-4pt}{\includegraphics[height=1.6em]{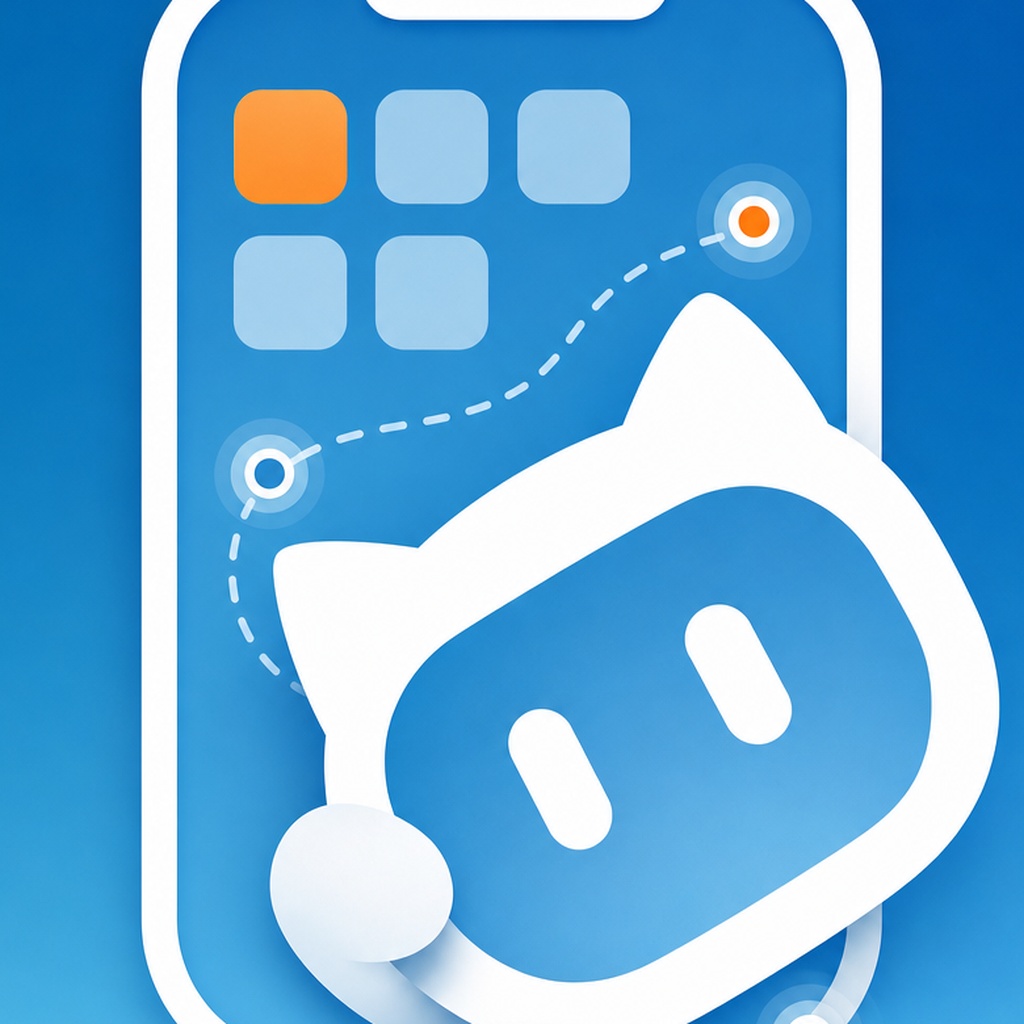}}~PhoneBuddy:\\[4pt]
Training Open Models for Agentic Phone Use\par}
\end{center}
\vskip 3mm
{\color{hunyuanblue}\hrule height 0.6pt}
\vskip 6mm
\begin{center}
\textbf{Zhengyang Tang}$^{1,2,*}$ \quad
\textbf{Xin Lai}$^{1,*}$ \quad
\textbf{Pengyuan Lyu}$^{1,*}$ \quad
\textbf{Xinyuan Wang}$^{1,*}$ \quad
\textbf{Tianyi Bai}$^{1,*}$ \\
\textbf{Chenxin Li}$^{1,*}$ \quad
\textbf{Yiduo Guo}$^{1,*}$ \quad
\textbf{Huawen Shen}$^{1,*}$ \quad
\textbf{Yuxuan Liu}$^{1,3,*}$ \\
\textbf{Junyi Li}$^{1}$ \quad
\textbf{Zhengyao Fang}$^{1}$ \quad
\textbf{Yang Ding}$^{1}$ \quad
\textbf{Yi Zhang}$^{1}$ \quad
\textbf{Weinong Wang}$^{1}$ \\
\textbf{Xingran Zhou}$^{1}$ \quad
\textbf{Liang Wu}$^{1}$ \quad
\textbf{Fei Tang}$^{1}$ \quad
\textbf{Sunqi Fan}$^{1}$ \quad
\textbf{Shangpin Peng}$^{1}$ \\
\textbf{Zheng Ruan}$^{1}$ \quad
\textbf{Anran Zhang}$^{1}$ \quad
\textbf{Benyou Wang}$^{2}$ \quad
\textbf{Ji-Rong Wen}$^{3}$ \quad
\textbf{Rui Yan}$^{4}$ \\
\textbf{Chengquan Zhang}$^{1,\dagger}$ \quad
\textbf{Han Hu}$^{1}$
\\[8pt]
$^1$Tencent Hunyuan \quad
$^2$The Chinese University of Hong Kong, Shenzhen \\
$^3$Gaoling School of Artificial Intelligence, Renmin University of China \quad
$^4$Wuhan University
\\[6pt]
{\small $^*$Equal contribution \quad $^\dagger$Project Lead \quad Correspondence to: \texttt{zhytang@tencent.com}}\\[3pt]
\end{center}
\vskip 4mm

\begin{tcolorbox}[
  colback=abstractbg, colframe=hunyuanblue, boxrule=0.6pt, arc=6pt,
  left=12pt, right=12pt, top=12pt, bottom=12pt]
\textbf{Abstract.}\quad
Phones are becoming an important execution surface for general-purpose agents, but training open models for reliable phone use remains difficult because the environment that matters at deployment, real devices running real apps, is slow, stateful, side-effectful, and hard to reset or verify, while scalable mock environments only approximate real behavior. We present \textit{PhoneBuddy}, a training recipe and open-model line for agentic phone use that combines a \textit{real-app environment} with a \textit{mock-app environment}, \textit{PhoneWorld}, which reconstructs runnable mock apps from real GUI usage structure. PhoneBuddy first builds a shared supervised fine-tuning stage from trajectories collected in both environments, then compares real-app RL against mixed RL across both environments. Across a 150-task human evaluation on real phones spanning apps, mini-apps, and cross-app workflows, task success rate improves from 36.67\% after supervised fine-tuning to 40.67\% after real-app RL and 45.33\% after mixed RL. On AndroidWorld, the same progression rises from 60.3\% to 77.2\% to 83.2\%. These results show that mock-app training is not a replacement for real-app RL, but a complementary source of scalable, resettable, and automatically checked interaction. The gains are strongest on app and mini-app tasks, while long-horizontal cross-app workflows remain an important open challenge.

\vskip 8pt
\textbf{Project Page:} \url{https://phonebuddyai.github.io/}

\vskip 4pt
\textbf{Date:} \today
\end{tcolorbox}

\section{Introduction}
\label{sec:introduction}

Large language models are increasingly expected not only to answer questions, but also to act through software interfaces. Recent work has pushed this direction across web agents, desktop operating-system agents, tool-using agents, and mobile GUI agents~\citep{deng2023mind2web,zhou2023webarena,koh2024visualwebarena,xie2024osworld,bonatti2024windowsarena,yang2025macosworld,zhang2023appagent,wang2024mobileagent,rawles2024androidworld}. Phones are a particularly important execution surface because they are the primary interface for messaging, payments, local services, mini-app ecosystems, personal data, and everyday cross-application workflows. A phone agent is therefore not useful merely because it can recognize widgets or describe a screen; it must reliably complete user tasks under real device state, real application behavior, and real user-facing side effects. This makes phone use a harder target than static GUI grounding: success depends on reading the current screen, deciding which action is safe and useful, maintaining progress over many steps, and verifying that the intended outcome actually happened.

The difficulty is amplified by the structure of real phone tasks. Mobile tasks are stateful, permission-rich, and side-effectful; they depend on login status, app-specific business logic, notification state, device settings, prior user data, and sometimes opaque server-side behavior. They also appear in several interaction regimes: single native apps, mini-apps embedded in host platforms, and cross-app workflows that require transferring information between interfaces. Existing datasets, benchmarks, and agents have made progress on mobile screen understanding, action prediction, real-device evaluation, and long-horizon interaction~\citep{rawles2023androidwild,deng2024mobilebench,wang2024mobileagentbench,xu2025mobilebenchv2,xu2025androidlab,kong2025mobileworld,chai2025a3,liu2025mobilesteward}. These advances are necessary, but they leave a narrower training question unresolved: \textbf{how should an open phone-use model be trained so that it improves task completion on real phones, rather than only improving local action imitation or benchmark-specific interaction?}

Our starting point is the mismatch between realism and scalability. A real-app environment, where agents operate authentic apps on real devices, is the setting that ultimately matters and exposes account-dependent behavior, real side effects, app instability, permission flows, and the gap between apparent progress and completed tasks. However, it is expensive to scale, hard to reset, and difficult to verify automatically. A mock-app environment can be reset, repeated, instrumented, and checked at much lower cost, but it risks training agents on simplified behavior that does not transfer to real phones. This tension appears broadly in recent work on synthetic environments, verifiable software worlds, GUI environment generation, and online RL for computer-use agents~\citep{zala2024envgen,cao2026guigenesis,dong2026agentworld,zhang2026infiniteweb,wu2026autowebworld,aggarwal2026gymanything,wang2026cuagym,wei2026opencomputer,lai2025computerrl,zhu2026workflowgym}. We argue that the practical training recipe should not choose one side of this tradeoff. Real-app training and mock-app training solve different parts of the same problem.

This paper introduces \textit{PhoneBuddy}, a training recipe and open-model line built around this complementarity. The real-app environment supplies realism and late-stage optimization on actual phone execution. The mock-app environment, \textit{PhoneWorld}, supplies scalable, resettable, and automatically verifiable interaction reconstructed from real GUI usage structure~\citep{tang2026phoneworld}. The central claim is not that PhoneWorld replaces real apps, or that real apps make mock apps unnecessary. Instead, the claim is that real-app RL and mock-app RL should be combined: real-app RL anchors the model to real device behavior and real side effects, while mock-app training adds broader and cheaper interaction signal from tasks that can be repeated and checked reliably. This framing also separates the training problem studied here from adjacent questions about runtime orchestration, privacy, and safety, which remain essential for deployable phone agents~\citep{phoneharness2026,tang2026privacy,phonesafety2026,deb2024agentdojo,levy2025safearena}.

Concretely, we study a compact open 4B model line under three stages: supervised fine-tuning, real-app RL, and mixed RL in both the real-app and mock-app environments. All compared checkpoints share the same Qwen3.5-4B backbone, action interface, and evaluation protocol; they differ only in the final training branch. On a 150-task human evaluation on real phones spanning apps, mini-apps, and cross-app workflows, task success rate improves from 36.67\% after supervised fine-tuning to 40.67\% after real-app RL and 45.33\% after adding mock-app training. On AndroidWorld, the same model line improves from 60.3\% to 77.2\% to 83.2\%. The gains are strongest on single-app and mini-app tasks, where workflow structure is stable and outcomes are easier to check, while cross-app workflows remain a major limitation. We view this boundary as part of the result: better training environments help substantially, but reliable phone agents still need stronger long-horizon state tracking, information handoff across apps, and runtime verification.

\textbf{Contributions.}
This paper makes the following contributions:

\noindent $\bullet$ We frame real-world agentic phone use as a training problem for open models, rather than only a GUI grounding problem.

\noindent $\bullet$ We present \textit{PhoneBuddy}, a training recipe that combines a real-app environment with \textit{PhoneWorld}, our mock-app environment built from real GUI usage structure.

\noindent $\bullet$ We show that the combination of real-app and mock-app RL produces stronger results than either supervised fine-tuning or real-app RL alone, improving task success rate from 36.67\% to 45.33\% on a 150-task real-phone human evaluation and from 60.3\% to 83.2\% on AndroidWorld.

\noindent $\bullet$ We clarify the current capability boundary of the approach: PhoneWorld-driven gains are strongest on app and mini-app tasks, while cross-app workflows remain a major open challenge for future training and system design.

\section{Background}
\label{sec:background}

\subsection{Mobile and GUI Agents}
\label{sec:background_gui_agents}

Recent GUI-agent research has moved from static screen understanding toward agents that can operate real software through visual observations, structured action spaces, and multi-step interaction. Web and desktop environments such as WebArena, VisualWebArena, OSWorld, Windows Agent Arena, and macOSWorld established that open-ended software tasks require grounding, planning, tool use, and robust execution rather than isolated perception~\citep{zhou2023webarena,koh2024visualwebarena,xie2024osworld,bonatti2024windowsarena,yang2025macosworld,huang2025mobileipl,liu2026come}. Tool-use and workflow benchmarks such as API-Bank, ToolLLM, tau-bench, WorkArena, Toolathlon, OSWorld-MCP, and CocoaBench further shifted evaluation toward executable tasks and outcome-based scoring~\citep{li2023apibank,qin2023toollm,yao2024taubench,drouin2024workarena,deng2024mobilebench,xu2025mobilebenchv2,li2025toolathlon,jia2025osworldmcp,cocoabench2026}. Mobile agents extend this challenge to smartphones, where touch actions, app navigation, permissions, account state, personal data, and embedded mini-app ecosystems become part of the task environment. Representative systems and datasets such as AppAgent, Mobile-Agent, Android in the Wild, AndroidWorld, MobileBench, AndroidLab, and MobileWorld have improved mobile action prediction, real-device evaluation, and long-horizon interaction~\citep{zhang2023appagent,wang2024mobileagent,rawles2023androidwild,rawles2024androidworld,deng2024mobilebench,xu2025androidlab,kong2025mobileworld,liu2025mobilesteward}. 
More recently, a series of GUI foundation models such as OS-Atlas, UI-TARS, UI-Venus, Step-GUI, GUI-Owl, and MAI-UI have substantially strengthened agent capabilities in screen understanding, task planning, and action execution, laying the groundwork for deploying GUI agents in real-world scenarios~\citep{wu2024osatlas,qin2025uitars,wang2025uitars2,gu2025uivenus,stepfun2025stepgui,ye2025guiowl,zhou2025mai}. These works motivate PhoneBuddy's focus on training models that can complete real phone tasks, not only predict plausible next actions.

\subsection{Environment Scaling and Online Optimization}
\label{sec:background_training_envs}

The central training bottleneck is environment scale. Real applications provide high-fidelity behavior, but collecting trajectories, resetting state, and verifying outcomes are expensive. Synthetic or reconstructed environments provide cheaper interaction and stronger supervision, but they must preserve enough structure to transfer to real software. Recent work on EnvGen, GUI-Genesis, Agent-World, InfiniteWeb, AutoWebWorld, Gym-Anything, CUA-Gym, OpenComputer, ComputerRL, and Workflow-GYM explores this broader direction of generated, verifiable, or online-trainable environments for agents~\citep{zala2024envgen,cao2026guigenesis,dong2026agentworld,zhang2026infiniteweb,wu2026autowebworld,aggarwal2026gymanything,wang2026cuagym,wei2026opencomputer,lai2025computerrl,chen2025step,zhu2026workflowgym,xu2026mobile}. PhoneWorld follows the same scaling logic in the phone domain: it reconstructs runnable mock apps from real GUI usage structure so that tasks can be reset, repeated, and checked automatically. PhoneBuddy asks how such mock-app training should be combined with real-app RL, rather than treating either environment as sufficient by itself.

\subsection{Agent Harness and Safety Protection}
\label{sec:background_safety}

Training improves the model policy, but a deployable agent also needs a harness that turns model predictions into controlled interaction with real software. Such a harness defines the observation stream, action schema, parser, execution backend, step budget, logging format, and task-level completion checks; it also decides when to use GUI actions, tool calls, CLI commands, or other execution channels. Recent work on tool and workflow agents shows that this runtime layer is part of the agent capability itself, especially when workflows evolve over time or require multiple interaction modes~\citep{li2025toolathlon,jia2025osworldmcp,yang2026clianything,li2026clawevallive}. PhoneHarness follows this direction for phone agents by coordinating mixed GUI, CLI, and MCP-style actions around a shared phone-task interface~\citep{phoneharness2026}. This paper focuses on the training recipe, but we treat the harness as a necessary deployment layer: it mediates between model outputs and real side effects, provides execution traces for debugging and learning, and supplies the structure needed for future online adaptation~\citep{huang2026towards}. Phone agents also operate close to sensitive user data, so the harness must act as a safety boundary rather than only an action executor. Prior work on sandboxed risk evaluation, web-agent safety, phone privacy, and phone safety shows that capable agents still require guardrails, explicit runtime boundaries, permission checks, and careful evaluation of harmful or privacy-sensitive behavior~\citep{ruan2023toolemu,deb2024agentdojo,zhang2024agentsafetybench,levy2025safearena,tang2026privacy,phonesafety2026}.

\section{Method}
\label{sec:method}

\subsection{Problem Setting}
\label{sec:problem_setting}

PhoneBuddy targets the final stage of training a phone-use model. To solve user instruction, at each step, the agent observes the current screen together with the interaction history, and predicts next action. An episode ends when the agent declares the task finished or exhausts its step budget.

The central difficulty stems from a mismatch between the requirements of training and deployment. An effective training environment should be easy to reset and to verify automatically, so that the policy can be optimized against reliable, repeatable outcome signals. The deployment target, however, is a real phone running authentic apps, whose persistent state and irreversible side effects are precisely what make resetting and automatic verification costly. The two demands therefore stand in tension, and neither can be satisfied by a single environment alone. PhoneBuddy is designed to bridge this gap by training across both: a real-app environment for fidelity and a mock-app environment for scalable, verifiable interaction.

\subsection{Overview of PhoneBuddy}
\label{sec:overview_phonebuddy}

\begin{figure*}[t]
\centering
\includegraphics[width=0.85\linewidth]{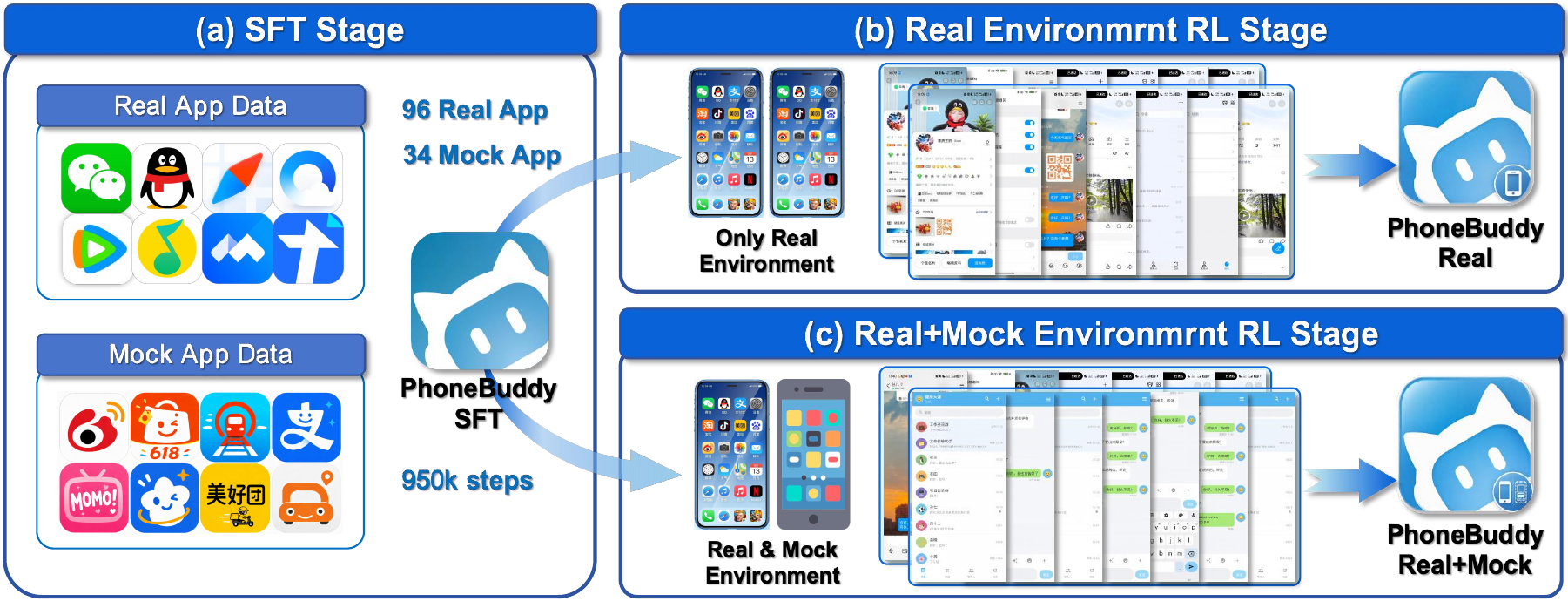}
\caption{Overview of PhoneBuddy. A shared SFT stage uses trajectories from both the real-app and mock-app environments, after which the same PhoneBuddy-4B-SFT model is branched into a real-app RL checkpoint and a mixed real+mock RL checkpoint. }
\label{fig:overview}
\end{figure*}

PhoneBuddy is a training recipe that turns a single base model into a phone-use agent through a shared supervised fine-tuning (SFT) stage followed by reinforcement learning across two complementary environments, as illustrated in Figure~\ref{fig:overview}. All checkpoints start from the same Qwen3.5-4B backbone and share the same SFT initialization, action interface, and evaluation protocol, differing only in the final RL branch. This design isolates our object of study---how the choice of reinforcement-learning environment affects the agent's ability to complete real phone tasks---so that any difference in task success is attributable to the final training stage alone.

Throughout the paper we distinguish two training environments:

\noindent $\bullet$ a \textbf{real-app environment}, in which the agent operates authentic apps on real devices;

\noindent $\bullet$ a \textbf{mock-app environment}, in which the agent operates runnable mock apps that can be reset and verified automatically. The environment used is \textbf{PhoneWorld}~\citep{tang2026phoneworld}.

Starting from the shared SFT checkpoint, we compare three variants: the SFT baseline itself (\textit{PhoneBuddy-4B-SFT}), a model further trained with RL only in the real-app environment (\textit{PhoneBuddy-4B-Real}), and a model further trained with mixed RL in both real-app and mock-app environments (\textit{PhoneBuddy-4B-Real+Mock}).

\subsection{Real-App Environment}
\label{sec:real_app_environment}

The real-app environment runs authentic apps on physical devices. It is indispensable because it is the environment the model must ultimately operate in: it faithfully exposes the real app behavior, device state, timing variation, and user-facing side effects that govern actual phone use.

Crucially, it surfaces failure modes that mock apps cannot fully reproduce, such as account-dependent behavior, app-specific instability, permission flows, and the gap between apparent progress and genuine task completion. It also enables \emph{real-app RL}, which we treat as the primary late-stage step for improving task completion on real phones.

Its drawback is cost: rollouts are slower, state is harder to reset, automatic verification is more fragile than in a mock-app environment, and exploration carries real, sometimes irreversible side effects that demand additional risk controls. PhoneBuddy therefore uses the real-app environment selectively, to keep training aligned with deployment while avoiding the cost of relying on it alone.

\begin{figure*}[t]
\centering
\includegraphics[width=\linewidth]{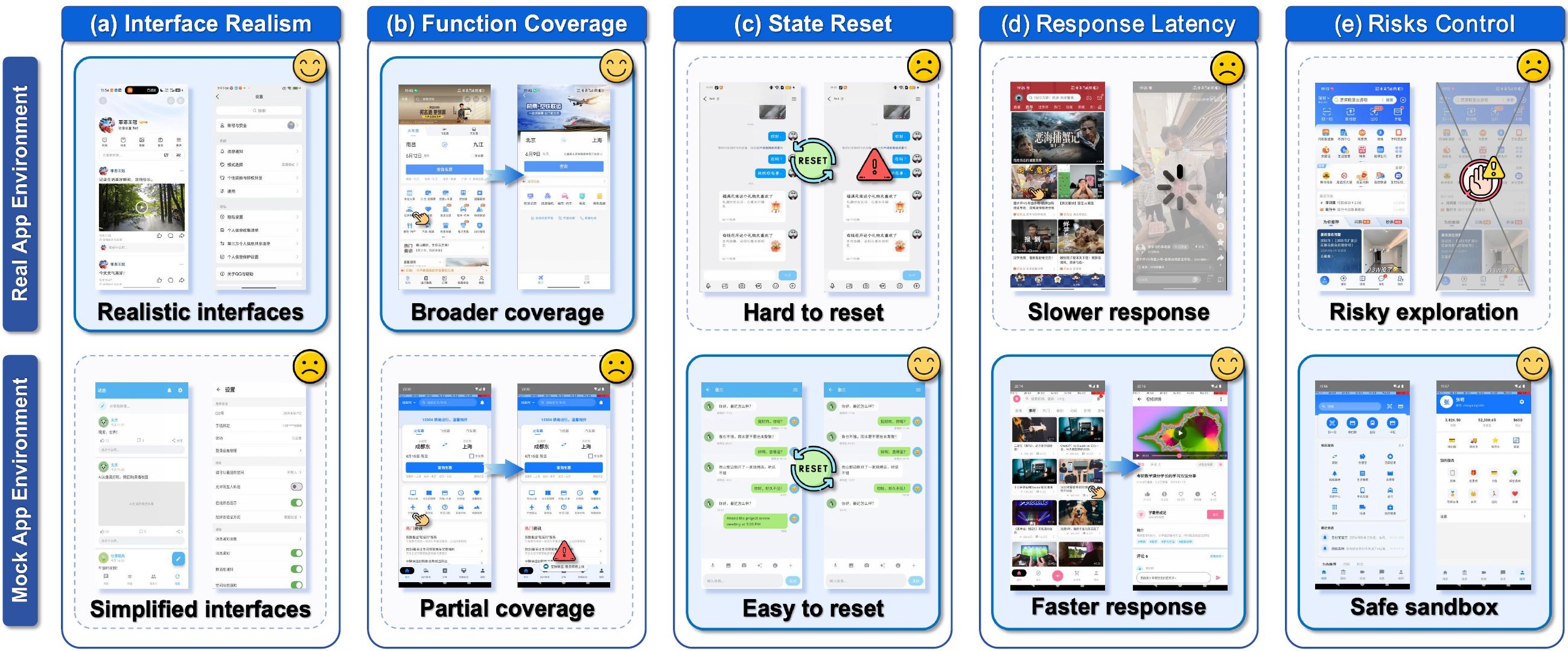}
\caption{Complementary roles of the real-app and mock-app environments. The real-app environment provides authentic device behavior, app logic, and user-facing side effects, while PhoneWorld provides resettable mock apps, automatic verification, and scalable rollout collection. PhoneBuddy uses both environments rather than treating either one as a complete substitute for the other.}
\label{fig:real_mock_comp}
\end{figure*}

\subsection{Mock-App Environment (PhoneWorld)}
\label{sec:method_phoneworld}

\textit{PhoneWorld}~\citep{tang2026phoneworld} is our mock-app environment. Here ``mock app'' does not mean a toy demo or a static prototype. It means a runnable Android app reconstructed from real GUI traces, with state that can change and with rules for checking whether a task is finished.

PhoneWorld employs a pipeline to build mock apps from real GUI trajectories and screenshots. From them, it recovers which screens matter, how screens connect, which actions need to be supported, and which state changes need to be saved. It then builds runnable mock Android apps with both read-only content and writable state. From the same apps, it derives tasks and rule-based verifiers so that success can be checked automatically rather than by manual inspection.

In its current version, PhoneWorld spans dozens of consumer-style mobile environments and supplies a large pool of executable tasks and trajectories. For the purposes of this paper, what matters is not the implementation details of any individual generated app, but the role PhoneWorld plays in training: it provides scale, repeatability, and automatic verification precisely in the setting where real-app training is most constrained.

\begin{table*}[t]
\centering
\small
\setlength{\tabcolsep}{5pt}
\begin{tabular}{p{3.0cm}p{3.0cm}p{2.4cm}p{3.3cm}p{3.0cm}}
\toprule
Checkpoint & Shared SFT Data & RL Environment & Training Objective & Purpose \\
\midrule
\textbf{PhoneBuddy-4B-SFT} & Real-app + mock-app trajectories & -- & Supervised fine-tuning & Common starting point before RL \\
\textbf{PhoneBuddy-4B-Real} & Real-app + mock-app trajectories & Real-app only & Reinforcement learning on real phone execution & Improve real-phone task completion \\
\textbf{PhoneBuddy-4B-Real+Mock} & Real-app + mock-app trajectories & Real-app + mock-app & Mixed reinforcement learning in both environments & Combine real execution with scalable verified interaction \\
\bottomrule
\end{tabular}
\caption{Training recipe used in the main comparison. All three checkpoints share the same SFT stage and differ only in the final training branch.}
\label{tab:recipe}
\end{table*}

\subsection{Training Recipe}
\label{sec:late_stage_training}

Our main empirical study isolates the effect of the final training recipe. All compared checkpoints share the same backbone, action interface, and evaluation protocol.

All three checkpoints share the same supervised fine-tuning stage. In the current training stack, we first collect phone-use trajectories from both the real-app environment and the mock-app environment, and use them to build a shared SFT dataset. Starting from this shared SFT model, we then branch into two RL settings: RL only in the real-app environment, and mixed RL in both the real-app and mock-app environments.

This shared SFT stage matters for the comparison. It puts both environments into the same training format: the model sees the task instruction and current phone screen, and predicts the next phone action. As a result, the later comparison between \textit{PhoneBuddy-4B-Real} and \textit{PhoneBuddy-4B-Real+Mock} isolates the value of the RL branch rather than differences in the basic action interface.

The shared SFT stage starts from Qwen3.5-4B and uses a combined dataset of 950,758 action steps collected from the real-app and mock-app environments. We perform full-parameter fine-tuning for 1,115 optimizer steps with batch size 512. Training uses packed 8,192-token sequences, where shorter examples are concatenated with attention masking between segments to improve utilization. The learning rate decays from $1\times10^{-5}$ to $1\times10^{-6}$. Because multiple short examples can be packed into one training sequence, the product of batch size and optimizer steps should not be read as a direct count of raw action steps.


For both RL branches, we run 50 online RL steps after the shared SFT stage. Both branches optimize the same binary task-completion objective, but the reward must be instantiated differently because observability differs across environments. In the real-app environment, many task outcomes depend on account-specific or proprietary server-side state that is not directly accessible from the device interface. We therefore use rubric-based model judging over the observable interaction trace and UI evidence as a proxy for task completion. Concretely, for each instruction, we first use Gemini-3.1-Pro-Preview to generate task-specific rubrics, then use Qwen3.5-122B-A10B to score the trajectory against each rubric item; a rollout is counted as successful only if every rubric item passes. In the mock-app environment, PhoneWorld exposes built-in rule-based verifiers over the reconstructed app state, enabling automatic completion checks without model judging. Both signals are normalized to the same binary reward for policy optimization.

\paragraph{PhoneBuddy-4B-SFT.}
This is the supervised fine-tuning baseline. It is trained on phone-use trajectories to establish a common task-completion starting point before online optimization. In the main result table, this checkpoint serves as the reference for measuring the gains from RL and PhoneWorld augmentation.

\paragraph{PhoneBuddy-4B-Real.}
This checkpoint continues training with reinforcement learning in the real-app environment. The goal of this stage is simple: improve performance on real phone execution, including real app behavior, real device state transitions, and real user-facing side effects. We run 50 online RL steps in the real-app environment only. The reward uses the rubric-based model judging described above as a proxy for whether the intended phone task was completed under real execution.

\paragraph{PhoneBuddy-4B-Real+Mock.}
This checkpoint uses mixed RL in both the real-app environment and the mock-app environment. The key idea is not to replace real-app training, but to supplement it with broader and easier-to-verify phone-use interaction. PhoneWorld contributes task environment that can be reset, repeated, and checked automatically, while real-app RL keeps the model tied to real execution. We also run 50 online RL steps in this branch, with a 50\%/50\% real/mock rollout mixture. The two environments share the same high-level optimization target, task completion, while differing in how completion is verified.

At a high level, real-app rollouts use rubric-based model judging over the observable trajectory as a proxy for task completion, while mock-app rollouts use the built-in rule-based verifiers provided by PhoneWorld. This keeps the optimization target aligned across environments even though the reward is instantiated differently in each one.

\begin{highlightbox}{Why compare these three checkpoints?}
  The objective of this comparison is not to establish the value of PhoneWorld or real-app RL in isolation, but \textbf{to determine whether combining the real-app environment and mock-app environment constitutes a more effective training recipe for phone use agents}. Accordingly, these three checkpoints differ only in their final RL stage, which allows any difference in task success to be attributed directly to the choice of RL environment.
\end{highlightbox}

\section{Experimental Setup}
\label{sec:experimental_setup}

\subsection{Benchmarks and Metrics}
\label{sec:benchmarks_metrics}

We evaluate PhoneBuddy on four task settings. The first three come from our real-phone human evaluation suite: \textbf{Single-App Tasks}, \textbf{Cross-App Tasks}, and \textbf{WeChat Mini-App Tasks}, with 50 tasks in each category for a total of 150 tasks. The fourth setting is \textbf{AndroidWorld}. For all four settings, we report \textit{task success rate}. In our real-phone suite, a task is counted as successful only when it is fully completed.

For the main table, we report one number for each of these four settings and an overall \textbf{Avg.} computed as the unweighted mean of the four columns. This presentation gives a cleaner view of where the model is strong and where it still fails. In particular, it prevents improvements on one subset from hiding weaknesses on another subset.

\begin{figure*}[t]
\centering
\includegraphics[width=\linewidth]{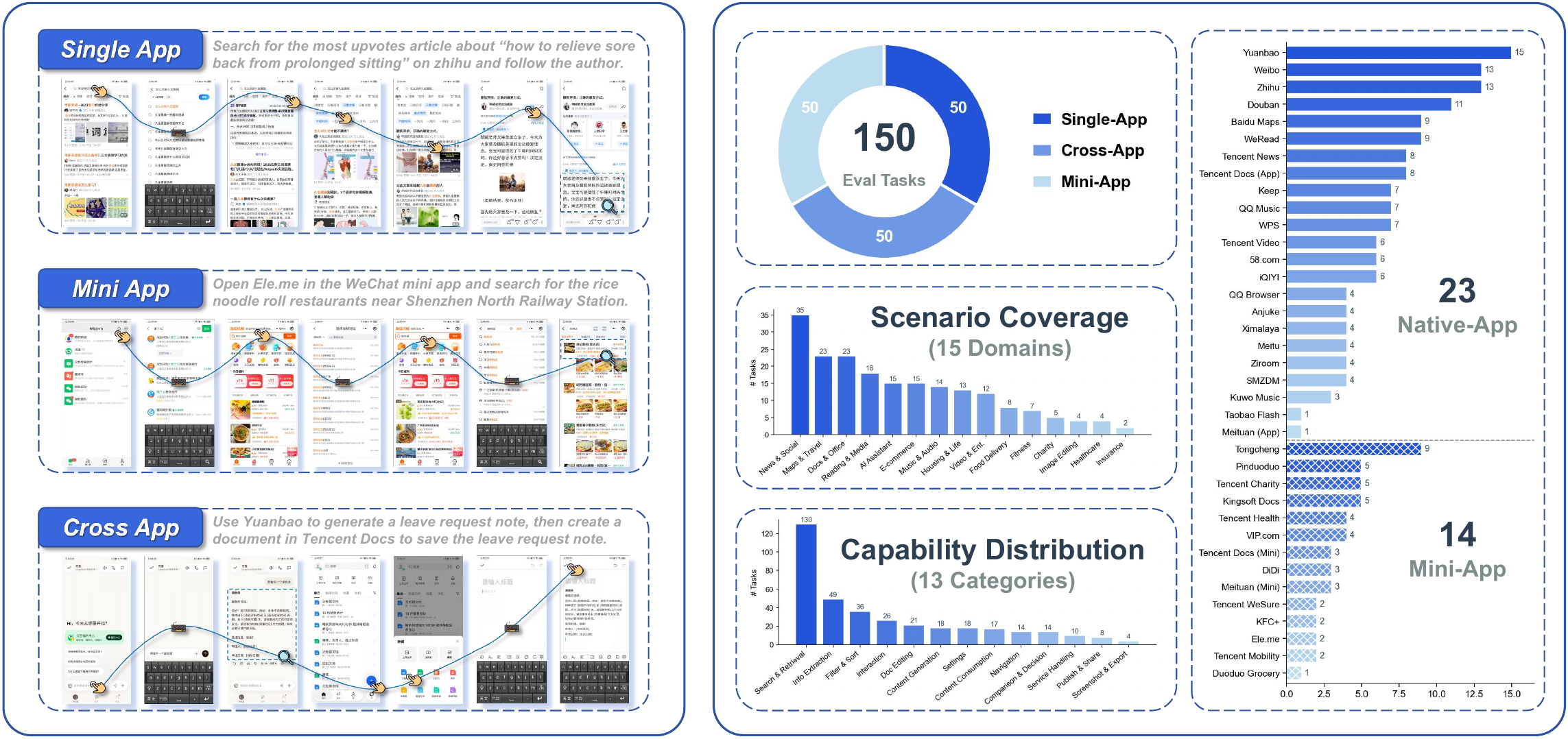}
\caption{Benchmark overview. The real-phone human evaluation covers Single-App, Cross-App, and WeChat Mini-App tasks, each with 50 tasks. All settings are evaluated with task success rate.}
\label{fig:benchmark}
\end{figure*}

\subsection{Evaluation Protocol}
\label{sec:evaluation_protocol}

We keep the action space, prompt template, step budget, and evaluation harness fixed across compared checkpoints, and change only the training recipe. All compared checkpoints are evaluated under the same inference and execution setup. For the real-phone human evaluation, a task is counted as successful if human annotators judge that the requested task has been fully completed on the device. We report task success rate only.

The action space is a shared phone-control API with normalized coordinates in the $[0,1000]$ range. The model predicts one action at each step from the following set: click, double click, long press, type, scroll, drag, button press with back/home/menu/enter, open app, close app, and wait. During training, the same prompt format also includes task-level communication actions for asking the user for clarification, outputting information, and marking the task as finished, but the core execution interface used for phone control remains fixed across compared checkpoints.

The prompt template is also held fixed. Each inference step is framed as a multimodal action prediction problem with the current screenshot and a structured textual context. The textual prompt contains the user instruction, a serialized history of prior thought-action pairs, and an intermediate-state field carried over from the previous step. The system-side prompt defines the full tool schema and instructs the model to output exactly one structured tool call enclosed by dedicated tags. The response may optionally include a reasoning block and an updated intermediate state, but execution uses only the parsed structured action. At inference time, we extract the tagged tool call, repair minor JSON formatting errors when needed, and map the result into a shared internal action representation for execution. This parsing layer is kept fixed for all compared models.

We use a maximum step budget of 30 during training and 50 during evaluation. The larger test-time budget reduces truncation on long-horizon tasks while preserving the same action interface, prompt contract, and execution stack across all compared checkpoints.

\subsection{Model Variants}
\label{sec:model_variants}

Our main internal comparison uses three checkpoints from the same 4B line:
\begin{itemize}
    \item \textbf{PhoneBuddy-4B-SFT}: the supervised fine-tuning baseline.
    \item \textbf{PhoneBuddy-4B-Real}: the model after real-app RL.
    \item \textbf{PhoneBuddy-4B-Real+Mock}: the model after mixed RL in both the real-app and mock-app environments.
\end{itemize}

We compare these models against representative strong closed-source systems, including Gemini 3.1 Pro, GPT-5.4, Claude Opus 4.7, and Seed 2.0 Pro.

\section{Main Results}
\label{sec:main_results}

Table~\ref{tab:main_results} reports task success rate across the four evaluation settings. We organize the results by setting, as this disaggregation reveals the central finding of our study: mock-app training yields consistent improvements over real-app RL on most settings, but the magnitude of this gain varies substantially across task types. We analyze each setting in turn below.

\begin{table*}[t]
\centering
\small
\setlength{\tabcolsep}{4pt}
\begin{tabular}{lccccc}
\toprule
Model & Single-App & Cross-App & WeChat Mini-App & AndroidWorld & Avg. \\
\midrule
Gemini 3.1 Pro & 50.0 & \textbf{48.0} & 58.0 & 80.2 & \textbf{59.1} \\
GPT-5.4 & 50.0 & 32.0 & 40.0 & 70.7 & 48.2 \\
Claude Opus 4.7 & 38.0 & 16.0 & 28.0 & 56.0 & 34.5 \\
Seed 2.0 Pro & 44.0 & 30.0 & \textbf{60.0} & 71.5 & 51.4 \\
\midrule
PhoneBuddy-4B-SFT & 34.0 & 22.0 & 54.0 & 60.3 & 42.6 \\
PhoneBuddy-4B-Real & 54.0 & 20.0 & 48.0 & 77.2 & 49.8 \\
PhoneBuddy-4B-Real+Mock & \textbf{62.0} & 18.0 & 56.0 & \textbf{83.2} & 54.8 \\
\bottomrule
\end{tabular}
\caption{Main results across four task settings. The first three columns come from the 150-task real-phone human evaluation, with 50 tasks each for Single-App, Cross-App, and WeChat Mini-App. All columns report task success rate. For the real-phone human evaluation, task success is defined strictly: a task counts as successful only when it is fully completed. Avg. is the unweighted mean of the four columns.}
\label{tab:main_results}
\end{table*}

\begin{figure*}[h]
\centering
\includegraphics[width=0.8\linewidth]{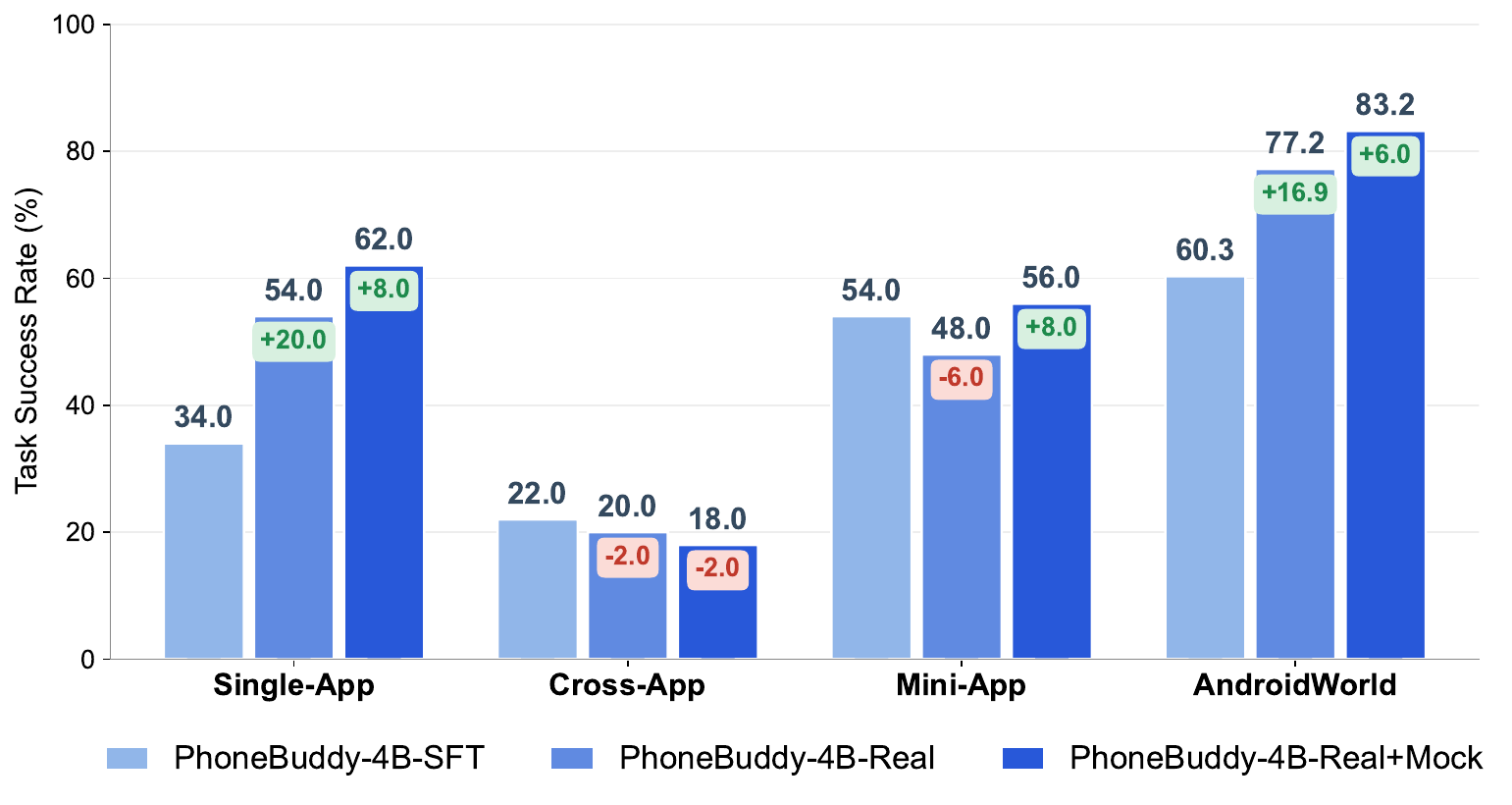}
\caption{Incremental gains from the two RL branches. The first delta measures the effect of real-app RL over the shared SFT checkpoint, and the second delta measures the additional effect of adding mock-app RL on top of real-app RL. The plot highlights that PhoneWorld improves Single-App, WeChat Mini-App, and AndroidWorld performance, while Cross-App tasks remain difficult.}
\label{fig:rl_delta}
\end{figure*}

\paragraph{Avg.}
The best overall internal model is \textit{PhoneBuddy-4B-Real+Mock}. It reaches 54.8 average task success rate across the four settings, improving over \textit{PhoneBuddy-4B-SFT} by 12.2 points and over \textit{PhoneBuddy-4B-Real} by 5.0 points. It also outperforms GPT-5.4 and Seed 2.0 Pro on this average, while remaining below Gemini 3.1 Pro overall.

\paragraph{Single-App Tasks.}
Single-App Tasks show the clearest gain from the full recipe. Performance rises from 34.0\% for \textit{PhoneBuddy-4B-SFT} to 54.0\% for \textit{PhoneBuddy-4B-Real}, and then to 62.0\% for \textit{PhoneBuddy-4B-Real+Mock}. This is the best performance on Single-App tasks, surpassing all compared closed models. The result suggests that real-app RL teaches the model to execute real phone actions more reliably, while mixed RL adds extra coverage on structured app interactions that benefit from repeatable training.

\paragraph{Cross-App Tasks.}
Cross-App Tasks remain the main gap. Performance stays low across all three checkpoints: 22.0\%, 20.0\%, and 18.0\% for \textit{PhoneBuddy-4B-SFT}, \textit{PhoneBuddy-4B-Real}, and \textit{PhoneBuddy-4B-Real+Mock}, respectively, so we do not observe a meaningful improvement from the current training recipe on this subset. A plausible explanation is task coverage. The current PhoneWorld task pool is primarily single-app, and the gains on WeChat mini-app tasks suggest that some of the learned interaction patterns can still transfer to mini-app settings. By contrast, cross-app workflows require explicit information handoff and persistent dependencies across multiple apps, which are not yet directly modeled in the current mock-app task pool. Extending PhoneWorld to cover such workflows is therefore an important direction for future work. Even with broader coverage, however, these tasks are likely to remain challenging because they also require stronger long-horizon state tracking, runtime coordination, and intermediate verification.

\paragraph{WeChat Mini-App Tasks.}
WeChat Mini-App Tasks show a different pattern. Real-app RL alone does not help this subset, dropping from 54.0\% to 48.0\%, but mixed RL lifts the score to 56.0\%. This is modestly above the SFT baseline and suggests that PhoneWorld is especially helpful when the workflow is multi-step but structurally stable, with state changes that are easier to verify and repeat during training.

\paragraph{AndroidWorld.}
AndroidWorld shows the cleanest monotonic trend: 60.3\% for \textit{PhoneBuddy-4B-SFT}, 77.2\% for \textit{PhoneBuddy-4B-Real}, and 83.2\% for \textit{PhoneBuddy-4B-Real+Mock}. The final model is also the best overall system in this column. This matters because AndroidWorld is outside the real-phone human evaluation suite used for the first three columns. The gain therefore supports the transfer value of the training recipe rather than only fitting to one internal benchmark.

\section{Qualitative Examples}
\label{sec:qualitative_examples}

\begin{figure*}[h]
    \centering
    \includegraphics[width=\linewidth]{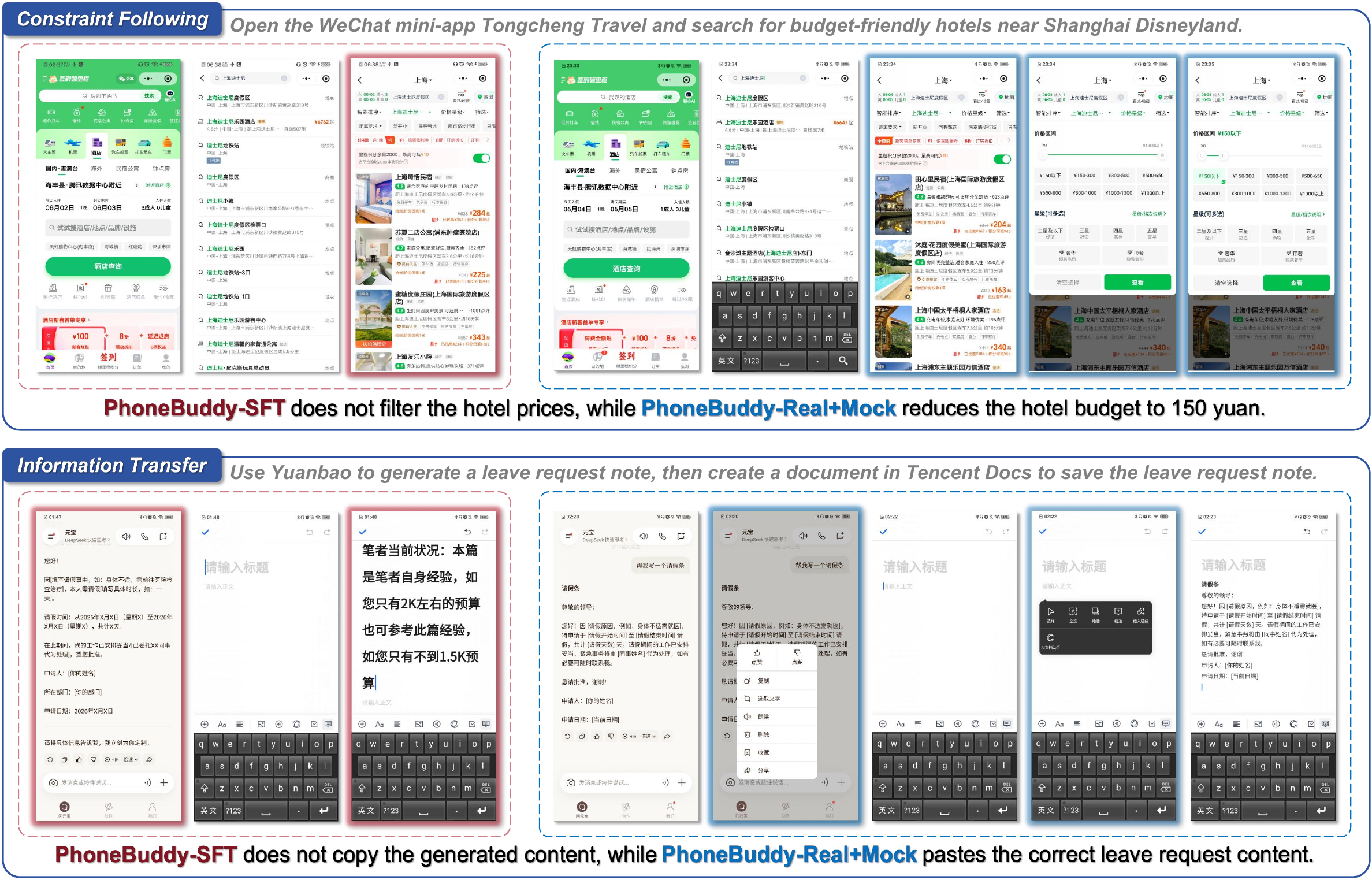}
    \caption{
    Representative successful trajectories. 
    PhoneBuddy-Real+Mock better preserves task constraints and information transfer.
    }
    \label{fig:success_cases}
\end{figure*}

Figure~\ref{fig:success_cases} shows two representative trajectories that reveal how Real+Mock training improves execution beyond the aggregate success rates. 

\noindent $\bullet$ In \textbf{constraint-following} case, the agent must search for budget-friendly hotels near Shanghai Disneyland in the WeChat mini-app Tongcheng Travel. PhoneBuddy-SFT reaches a plausible hotel-search page but does not apply the budget constraint, while PhoneBuddy-Real+Mock continues to the filtering interface and reduces the hotel budget to 150 yuan. 

\noindent $\bullet$ In \textbf{information-transfer} case, the agent must generate a leave request note with Yuanbao and save it in Tencent Docs. PhoneBuddy-SFT fails to copy the note generated by Yuanbao and instead inserts stale clipboard content when creating the document. By contrast, PhoneBuddy-Real+Mock correctly copies the generated leave request note and pastes it into the newly created document.

These cases suggest that mixed-environment RL does more than encourage broader exploration: it also supplies useful supervision for following task constraints and for the complex operations involved in transferring information.

\section{Discussion and Limitations}
\label{sec:discussion}

\paragraph{Why Real+Mock Works.}
The current results support a fairly specific conclusion. Real-app RL and PhoneWorld are complementary. Real-app RL ties the model to real device behavior, real app logic, and real side effects. PhoneWorld then adds scale, easier reset, and automatic verification. This combination is especially effective on tasks where the workflow is stable and the end state is easy to check.

\paragraph{Why Cross-App Still Lags.}
Cross-app execution remains a clear weakness. A likely factor is task coverage: the current PhoneWorld task pool is primarily single-app, although some of the resulting interaction patterns appear to transfer to mini-app settings. It does not yet provide direct support for multi-app information handoff, artifact transfer, or persistent cross-app state dependencies. Future work should extend mock environments to explicitly model these workflows. At the same time, cross-app tasks also stress long-horizon memory and runtime coordination, so broader environment coverage alone may not be sufficient.

\paragraph{What This Paper Does Not Solve.}
This paper is intentionally about training. A deployable phone agent also needs a strong runtime system and clear deployment boundaries around privacy and safety. Those pieces matter for real use, but they are deliberately not the empirical center of this report.

More broadly, PhoneBuddy is the training layer in a larger phone-agent matrix from our research line. PhoneWorld builds the mock-app environments used for scalable training and evaluation~\citep{tang2026phoneworld}. PhoneBuddy studies how to train the model itself. PhoneHarness studies runtime execution~\citep{phoneharness2026}, and PhonePrivacy / PhoneSafety study deployment boundaries~\citep{tang2026privacy,phonesafety2026}. This paper focuses only on the training layer, but it fits into that larger stack rather than standing alone.

\section{Conclusion}
\label{sec:conclusion}

This paper studies how to train open models for real-world agentic phone use. The main lesson is simple: real-app training alone is not enough, and mock-app training alone is not enough. Real-app RL provides realism; PhoneWorld provides scale, reset, and verification. In the current study, the strongest recipe is a shared SFT stage built from both environments followed by mixed RL across both environments. This recipe improves task success on both our real-phone human evaluation and AndroidWorld, supporting the view that mock-app interaction can transfer when it is grounded in realistic GUI structure. At the same time, the weak cross-app results show that environment scaling does not by itself solve long-horizon state tracking, information handoff, or runtime coordination. Future work should therefore combine better training environments with stronger execution harnesses, intermediate verification, and safety-aware deployment boundaries for real phone agents.

\bibliography{iclr2026_conference}

\end{document}